\documentclass[lettersize,journal]{IEEEtran}
\usepackage{amsmath,amsfonts}
\usepackage{algorithmic}
\usepackage{algorithm}
\usepackage{array}
\usepackage[caption=false,font=scriptsize,labelfont=sf,textfont=sf]{subfig}
\usepackage{textcomp}
\usepackage{stfloats}
\usepackage{url}
\usepackage{verbatim}
\usepackage{graphicx}
\usepackage{cite}
\usepackage{multirow}
\usepackage{booktabs}
\usepackage{hyperref}
\usepackage{xcolor}

\newcommand{\leftcell}[2][l]{%
  \begin{tabular}[#1]{@{}l@{}}#2\end{tabular}}
\newcommand{\specialcell}[2][c]{%
  \begin{tabular}[#1]{@{}c@{}}#2\end{tabular}}
\begin{document}

\title{COLA: COarse-LAbel multi-source LiDAR semantic segmentation for autonomous driving}

\author{Jules Sanchez$^1$, Jean-Emmanuel Deschaud$^1$, François Goulette$^{1,2}$
\thanks{$^1$Centre of Robotics, Mines Paris - PSL, PSL University, Paris, France. firstname.surname@minesparis.psl.eu}
\thanks{$^2$U2IS, ENSTA Paris, Institut Polytechnique de Paris, Palaiseau, France. firstname.surname@ensta-paris.fr}}



\maketitle

\begin{abstract}

LiDAR semantic segmentation for autonomous driving has been a growing field of interest in recent years. Datasets and methods have appeared and expanded very quickly, but methods have not been updated to exploit this new data availability and rely on the same classical datasets. Different ways of performing LIDAR semantic segmentation training and inference can be divided into several subfields, which include the following: domain generalization, source-to-source segmentation, and pre-training. In this work, we aim to improve results in all of these subfields with the novel approach of multi-source training. Multi-source training relies on the availability of various datasets at training time. To overcome the common obstacles in multi-source training, we introduce the coarse labels and call the newly created multi-source dataset COLA. We propose three applications of this new dataset that display systematic improvement over single-source strategies: COLA-DG for domain generalization (+10\%), COLA-S2S for source-to-source segmentation (+5.3\%), and COLA-PT for pre-training (+12\%). We demonstrate that multi-source approaches bring systematic improvement over single-source approaches.

\end{abstract}

\begin{IEEEkeywords}
Semantic Scene Understanding, Computer Vision for Transportation, Deep Learning in Robotics and Automation, 3D Computer Vision
\end{IEEEkeywords}

\section{Introduction}
\IEEEPARstart{T}{ransferability} and robustness have become the attention center for novel LiDAR semantic segmentation works. Specifically, domain adaptation, domain generalization, and pre-training are all of interest to re-usable models and training methods to improve performance and robustness.

Current works study the impact of multi-task training~\cite{lidog}, the use of the underlying geometry~\cite{3dlabelprop}, data augmentation strategies~\cite{lehner20233d}, and the availability of a large dataset for unsupervised training~\cite{segcontrast}. However, these works have only looked at single-source strategies, in which only one dataset is used at training time, and have left out multi-source strategies.

Multi-source training relies on the availability of various datasets to be used at training time. Implicitly, it is expected that these datasets represent a variety of domains, which means that some domain shifts exist between them. In the case of LiDAR scene understanding for autonomous driving, domain shifts can manifest themselves as a shift in the following: the acquisition hardware, such as sensor resolution; the geographical location of the acquired scenes; or scene type, such as urban centers or suburban cities. This variety is expected to improve the performances of the trained models. Following Sanchez~\textit{et al.}~\cite{3dlabelprop}, we will divide the domain shifts between sensor shift, scene shift, and appearance shift.

Multi-source strategies have been under-exploited in 3D scene understanding, with very few works looking into them~\cite{mdt3d,mantra,cola}. As pointed out in these works, this lack of interest stems from the difficulty of leveraging several datasets simultaneously due to their label sets' disparities. Some of the approaches in 2D scene understanding overcame this by re-annotating data~\cite{9157628}, which results in a significant human cost.

In Gao~\textit{et al.}~\cite{gao2021we}, the authors proposed an in-depth analysis of the currently available datasets for LiDAR semantic segmentation. They highlighted the various scenes, classes, and sensors for the currently available datasets. They concluded that current 3D deep learning approaches are hungry for diversity and size but that "due to the large  {domain gap}, mixing multiple datasets
in training may not improve model accuracy". {Contrary to their conclusion, we argue that multi-source semantic segmentation on LiDAR data, even with a significant domain gap between datasets, such as in outdoor environments, can enhance model performance.}

In our previous work~\cite{cola}, which this paper is based on, we proposed COLA, a relabelling strategy that allowed us to use several datasets simultaneously. This relabelling strategy relies on remapping existing labels to a common coarser set, which is an almost cost-free process. We employed this strategy for pre-training purposes only. In~\cite{cola}, COLA leveraged multi-source semantic segmentation as a pre-training task for outdoor LiDAR semantic segmentation. To achieve this, it presented a novel label set: the coarse labels specifically designed for LiDAR semantic segmentation for autonomous driving.

\begin{figure*}
    \centering
    \includegraphics[width=0.8\linewidth]{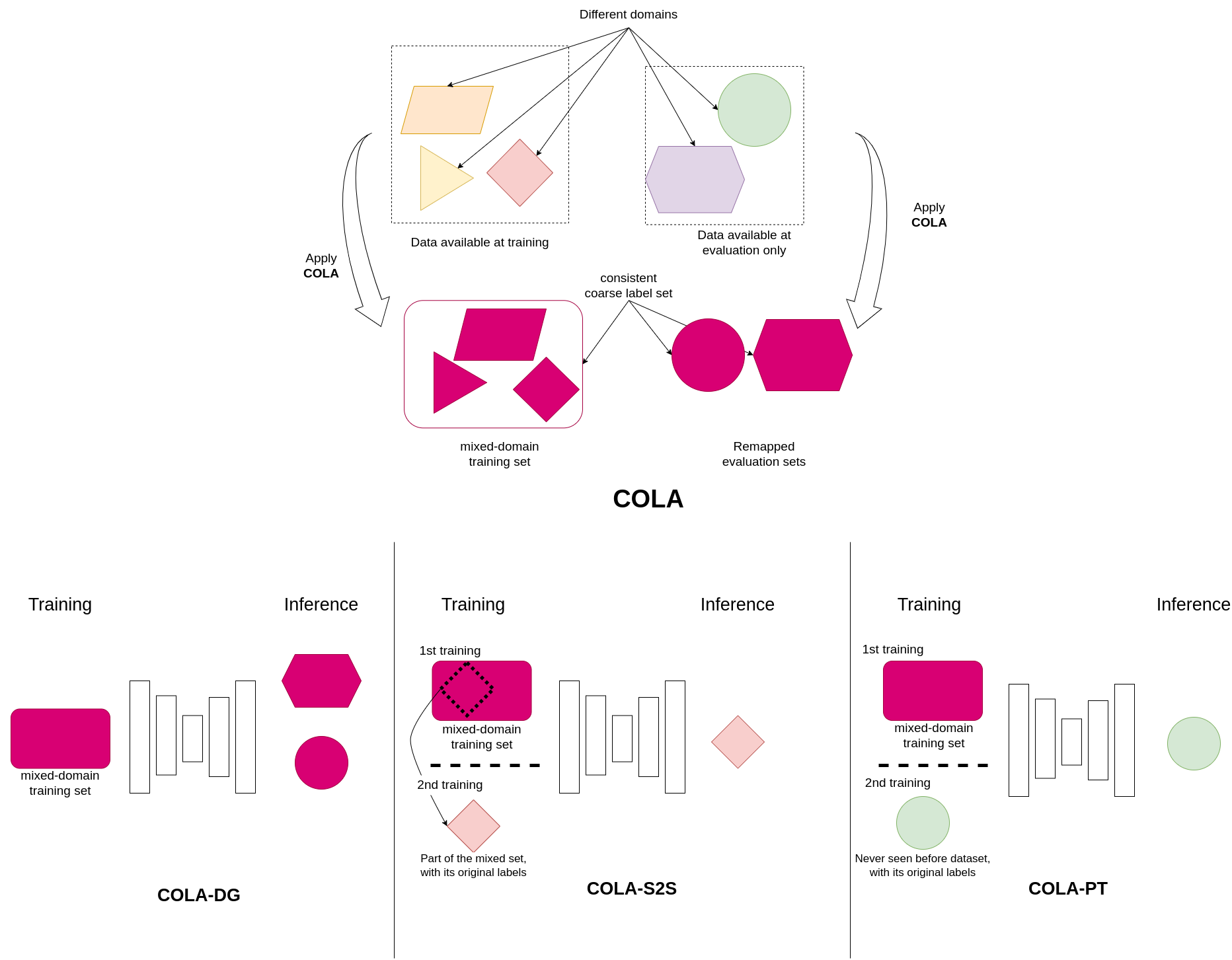}
    \caption{COLA and its applications: domain generalization (COLA-DG), source-to-source segmentation (COLA-S2S), and pre-training (COLA-PT). COLA-DG uses the mixed-domain training set to improve robustness and performance over unseen domains. COLA-S2S leverages the mixed-domain training set as a basis and complements it with a sample of the target set. COLA-PT uses the mixed-domain training set to extract a pre-trained model that can be finetuned on any target set.}
    \label{fig:COLA_new}
\end{figure*}

In this work, we expand COLA to several other cases, namely domain generalization and source-to-source semantic segmentation. Furthermore, we update and enrich experiments conducted in~\cite{cola} for pre-training. The difference between these three tasks can be found in \autoref{fig:COLA_new}. Contrary to~\cite{cola}, all experiments here are done on at least two largely different architectures to ensure that the conclusions are meaningful and done on a wider variety of data for pre-training and evaluation.

Overall, the goal of this work is to highlight the effectiveness of multi-source strategies, and we can summarize its contributions as follows:

\begin{itemize}
    \item We re-introduce COLA, a relabelling strategy that allows for the easy implementation of multi-source training for LiDAR semantic segmentation for autonomous driving.
    \item We propose the first multi-source domain generalization method, COLA-DG, for LiDAR semantic segmentation for autonomous driving.
    \item We propose a multi-source semantic segmentation strategy to improve source-to-source performances, called COLA-S2S.
    \item We update the study of multi-source pre-training, called COLA-PT, by comparing it with more recent works and a wider variety of models.
\end{itemize}

\section{Related work}

\subsection{LiDAR semantic segmentation}

LiDAR semantic segmentation (LSS) for autonomous driving requires high-speed performance of at least 10 frames processed per second. As such, older 3D deep learning relying on redefining convolutions, such as TangentConv~\cite{Tat2018}, SpiderCNN~\cite{10.1007/978-3-030-01237-3_6}, and KPConv~\cite{thomas2019KPConv}, are not used despite reaching satisfying performances. Methods that use structured representations are favored, and they include the following: projecting the point clouds in 2D with a range-based image~\cite{8967762,squeezesegv3}, bird's-eye-view image~\cite{Zhang_2020_CVPR} and leveraging sparse convolution-based architectures as introduced by MinkowskiNet~\cite{mink} under the name Sparse Residual U-Net (SRU-Net). SRU-Nets have been used as the backbone for more refined methods, such as point-voxel methods~\cite{spvnas}, cylindrical voxel convolutions as done by Cylinder3D~\cite{Zhou2020Cylinder3DAE}, and mixed-input representation methods~\cite{Xu2021RPVNetAD}. These methods are the best performing.

\subsection{From domain adaptation to domain generalization for LSS}

As mentioned in the introduction, the focus has shifted from source-to-source performances, in which the training and evaluation sets belong to the same domain, to robustness evaluation. Domain generalization addresses the transfer ability of a model or an algorithm from one source to a target domain without the availability of the target domain before the inference. Domain adaptation relies on a few samples from the target domain to improve its performance. 

Domain adaptation methods that have had the most traction recently are unsupervised domain adaptation (UDA) methods. { They assume that no annotations are available for the target domain.} Among them GIPSO~\cite{saltori2022gipso}, COSMIX~\cite{saltori2023compositional}, xMUDA~\cite{jaritz2020xmuda} and SALUDA~\cite{michele2024saluda} are some of the most popular. Most of these approaches use pseudo-labeling of the target scene to enhance the model's performances on the new domain; their differences lie in the strategy to ensure the quality of their pseudo-label.

In practice, it is often possible to have at least a few annotations from the target domain. Supervised domain adaptation has been almost not explored for 3D applications. SSDA3D~\cite{wang2023ssda3d} tackles this subject for object detection thanks to domain mixing.

\cite{generalizationsurvey,generalizationsurvey2} proposes a survey of domain generalization and a classification of the various strategies adopted. Among these methods, we can cite meta-learning~\cite{maml,episodic}, multi-task training~\cite{multitask,jigsaw}, data augmentation~\cite{adversarial}, neural network design~\cite{instancenorm,stylenorm,ibnnet}, and domain alignment~\cite{mixture,distributionmatching}.

These methods and paradigms were designed for 2D computer vision; more recently, researchers have created methods specifically designed for 3D. Among them, we can cite  DGLSS~\cite{domgen}, LIDOG~\cite{lidog}, 3D-VField~\cite{lehner20233d} and 3DLabelProp~\cite{3dlabelprop}. 

3D-VField~\cite{lehner20233d} is an adversarial data augmentation method that relies on adversarially attacking scene objects to robustify predictions. DGLSS~\cite{domgen} is a domain alignment method that uses a degraded version of the considered scan at training time to reduce the dependency to the resolution of the sensors, 3DLabelProp~\cite{3dlabelprop} is a domain alignment method as well which uses pseudo-dense point clouds to increase the geometric similarity between point clouds acquired with various sensors. Finally, LIDOG~\cite{lidog} is a multi-task method that segments the point clouds both in the voxel space and in the bird's-eye-view space.

All these methods are single-source, meaning they use only one dataset at training time despite the growing number of datasets available.

\subsection{Multi-source computer vision}

Multi-source training has been used to improve the robustness of computer vision deep learning models. When all the datasets have the same labels, it is easy to use, as simply concatenating datasets together is already a multi-source strategy and improves robustness. More elaborate strategies exist, such as meta-learning~\cite{maml}. Li~\textit{et al.}~\cite{episodic} leverages the availability of several datasets to ensure good domain generalization performance.

In most cases, available datasets do not have the same label set, and intermediate steps must be taken to carry out multi-source training. The most naive approaches propose a new label set from existing ones by taking the intersection~\cite{Ros2016TrainingCD,9707049}, union~\cite{10.1007/978-3-030-58568-6_11}, or sum with multi-head (MH) training~\cite{10.1007/978-3-030-30645-8_28,zhou2022simple}, which has one classification head per dataset in the training set. 

These methods strongly assume the inference label set. In the case of domain generalization, the target label set cannot be guaranteed to lie in the union of the training set, making the union and MH methods unreliable. The intersection methods can result in many interesting labels being discarded and depend on annotation choices.

Other works observed that despite having different label sets, datasets designed for similar tasks share ontological connections between their label sets. Some works have tried to explicitly model these relationships~\cite{8659045,8500398,8578183}. In 3D, Gao~\textit{et al.}~\cite{gao2021we} highlighted label hierarchies between SemanticKITTI and SemanticPOSS that showed that, while they have different label sets, { they share a similar set of classes.}
 
Finally, Liang~\textit{et al.}~\cite{mantra} noticed the variety of 3D datasets and the complexity of using them together. To overcome this, they used a pre-trained language model to encode label proximity between label sets. Through this, they could make predictions in a continuous label space rather than in a discrete one. Liang~\textit{et al.}~\cite{mantra} applied this work only to indoor scene understanding.

\subsection{Label efficiency for LSS}

Label efficiency for LiDAR semantic segmentation encompasses all methods that use as little annotations as possible to reach { competitive performance} compared to methods leveraging dense regular annotations. The main approach for label efficiency for LiDAR is pre-training, which is the ability to create models that understand the underlying geometry of the scenes which makes them re-employable for various tasks.

There are several strategies for pre-training, such as performing auxiliary tasks on the same data as the finetuning one or performing a simple task on a large quantity of data beforehand to learn general representations. As there are no very large annotated data for 3D, unsupervised methods are largely favored, especially contrastive training.

PointContrast~\cite{xie2020pointcontrast}, CSC~\cite{csc2021}, and DepthContrast~\cite{Zhang_2021_ICCV} are common methods for contrastive
learning. The differences between these methods stem from the modality of the input data, point clouds or RGB-D images, or the tweaks on the loss function. PC-FractalDB~\cite{Yamada_2022_CVPR} proposes an alternative to typical contrastive approaches by learning fractal geometry as a pre-training. These methods are particularly efficient for indoor scene understanding as there is more available data for this task \cite{dai2017scannet,shapenet2015,7298801}.

Two methods were specifically designed for LSS pre-training: SegContrast~\cite{segcontrast} and TARL~\cite{nunes2023temporal}. SegContrast is inspired by PointContrast but generates positive pairs by data augmentation rather than registration. TARL leverages a vehicle model to create object pairs based on the acquisition trajectory and tries to minimize the distance between their representation in two different point clouds. In both cases, it is unsupervised training, which they perform on SemanticKITTI~\cite{behley2019iccv}.

Other work tackled label efficiency without pre-training by smartly leveraging the few annotations available to improve performance, such as COARSE3D~\cite{li2022coarse3d} which uses contrastive learning and entropy-based sampling of the annotations, and Scribble supervision~\cite{unal2022scribble}.

Our work is at the intersection of these different fields. We propose a novel method of performing multi-source learning for LSS that can improve source-to-source or domain generalization performance or even be used for training by enabling the leverage of large annotated datasets. 

\section{COLA}

\subsection{Datasets}

\begin{table*}[h]
    \centering
    \small
    \begin{tabular}{l||cccllll}
        \multicolumn{1}{c}{Name} & \# scans  & \# sequences & \# labels &  Scene & Country & Sensor &  Manufacturer  \\\midrule \hline
        SemanticKITTI \cite{behley2019iccv}&23000 & 11 & 19 & Suburban & Germany & HDL-64E & Velodyne\\ \hline
        KITTI-360 \cite{kitti360}& 67626 & 9 & 18 & Suburban & Germany & HDL-64E & Velodyne\\ \hline
        nuScenes \cite{nuscenes}&35000 & 850 & 16 & Urban & \leftcell{Singapore\\USA} & HDL-32E & Velodyne\\ \hline
        Waymo \cite{waymo}& 30000 & 1150 & 22 & \leftcell{Suburban\\Urban} & USA& \leftcell{N/C \\ 64 beams} & Undisclosed \\ \hline \hline
        SemanticPOSS \cite{semanticposs}&3000 & 6 &13  & Campus & China& Pandora &Hesai\\ \hline
        PandaSet \cite{pandaset}& 6000& 76&36 & \leftcell{Suburban\\Urban} & USA & \leftcell{Pandar64 \\ PandarGT} & Hesai \\ \hline 
        ParisLuco3D \cite{sanchez2023parisluco3d}& 7500& 1& 45 & \leftcell{Urban} & France & \leftcell{HDL-32E} & Velodyne \\ \hline \midrule
    \end{tabular}
    \caption{Summary of the various datasets properties. Number of scans and number of sequences reflect the available annotated data.}
    \label{tab:summary_dataset}
\end{table*}

Several datasets need to be available to perform multi-source training. For this work, we use seven different datasets that we split into two categories: training sets (SemanticKITTI~\cite{behley2019iccv}, KITTI-360~\cite{kitti360}, nuScenes~\cite{nuscenes}, and Waymo~\cite{waymo}) and evaluation sets (SemanticPOSS~\cite{semanticposs}, PandaSet~\cite{pandaset}, ParisLuco3D~\cite{sanchez2023parisluco3d}).

PandaSet is divided into two subsets, Panda64 and PandaFF. This partition is made by separating the scans according to their acquisition sensors. Panda64 is made of scans acquired by the 64-beam LiDAR and PandaFF by a solid-state LiDAR. Thus, both datasets present the same scene but with very different acquisition sensors.

Training sets are selected due to their size, with more than 20,000 scans dedicated to training. Furthermore, they represent three different acquisition sensors and three different acquisition settings, providing a variety of domains for training, and thus can be safely considered a multi-source dataset. Details about the datasets can be found in \autoref{tab:summary_dataset}. Among them, SemanticKITTI and nuScenes are considered the standard datasets for LSS and will be used for comparisons.

Evaluation sets are selected due to their complexity and difference from the training sets. To verify the generalization capabilities of LSS methods, a significant domain gap needs to be found between training and evaluation sets. PandaFF, with its solid-state LiDAR, provides a strong sensor shift, whereas SemanticPOSS and ParisLuco3D provide strong scene shifts by being acquired in densely populated scenes. The three are small datasets, but they are on par with sequence 08 from SemanticKITTI, which is widely used for method comparison. Details about the datasets can be found in \autoref{tab:summary_dataset}.

\subsection{Current Limitations}

As highlighted by the~\autoref{tab:summary_dataset}, every LSS dataset has its own set of labels, and no one-to-one correspondence exists between any datasets. Like the issue highlighted by Liang~\textit{et al.}~\cite{mantra}, LSS datasets cannot be concatenated to create a multi-source training set. This is not only an issue for training but also evaluation. 

\begin{figure}[h]
    \centering
    \includegraphics[width=\linewidth]{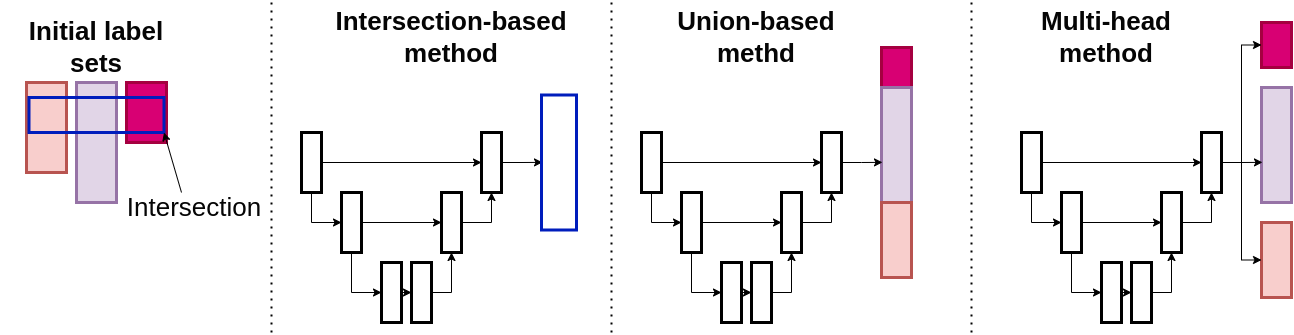}
    \caption{Illustration of naive multi-source methods.}
    \label{fig:ms_semantic_segmentation}
\end{figure}

As mentioned in the related work section, applying union or intersection to create a new label set requires strong constraints on the evaluation set, which are not fulfilled here, in \autoref{fig:ms_semantic_segmentation} a reminder of the naive approach for multi-source segmentation. Inspecting the label sets reveals how the union would fail, as some evaluation sets display very fine annotations, namely PandaSet and ParisLuco3D. Furthermore, the union strategy amplifies the label imbalance observed in the autonomous driving datasets.

Regarding the intersection, we observe that it raises the issue of large amounts of information being discarded even for the training set, as label sets do not agree on a common definition of buildings.


Other 3D domain generalization methods provided label mapping, which combined label sets and group labels under a common coarser label to manage cross-domain evaluation. Kim~\textit{et al.}~\cite{domgen} found a common set for SemanticKITTI, Waymo, nuScenes, and SemanticPOSS, but it has flaws. Some labels are discarded in a background label, and SemanticPOSS cannot be easily mapped to this label set. Sanchez~\textit{et al.}~\cite{3dlabelprop} needs to define a novel label set for each pair of datasets, making it especially tedious to use. 

\subsection{Coarse labels and their use}

Label sets can sometimes be represented as label trees, as is done by CityScapes~\cite{cordts2016cityscapes}, where they { differentiate categories from classes}.
Following these representations, we similarly introduce coarse labels, which are common in datasets annotated for similar tasks, and fine labels, which are the ones typically used for semantic segmentation. Given sufficiently coarse labels, they can be applied to any autonomous driving label set for almost no human cost by extracting this tree-like structure and merging labels to make these \textbf{coarse labels}. Such tree-like structures to represent label sets have also been used to compare label sets by Gao~\textit{et al.}~\cite{gao2021we}, but they did not attempt to apply them to a large variety of datasets.

\begin{figure*}[h]
    \centering
    \includegraphics[width=\linewidth]{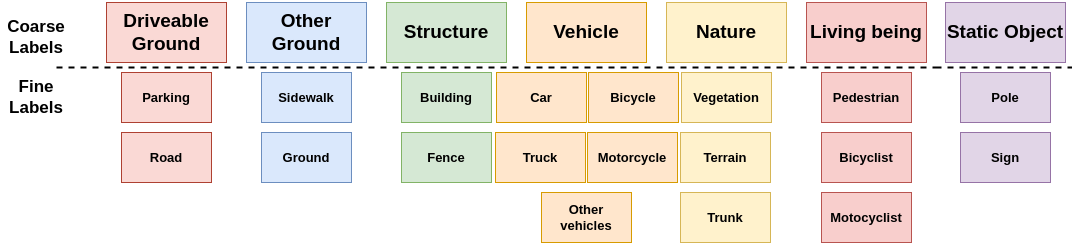}
    \caption{SemanticKITTI's label set and mapping to the coarse labels.}
    \label{fig:Hierarchie}
\end{figure*}

We identify seven unambiguous labels that can be used for any autonomous driving dataset: \textit{Vehicle}, \textit{Driveable Ground}, \textit{Other Ground}, \textit{Person}, \textit{Object}, \textit{Structure}, and \textit{Vegetation}. The process of remapping to them is shown in~\autoref{fig:Hierarchie}, in which SemanticKITTI's label set is remapped to the coarse labels. The mapping from each dataset is available in \autoref{tab:coarselabelsex}. It will also be uploaded to Github.

\begin{table*}[]
    \scriptsize
    \begin{tabular}{>{}c||>{}c|>{}c|>{}c|>{}c|>{}c|>{}c|>{}c}
         & \multicolumn{1}{>{}c}{Driveable Ground} & \multicolumn{1}{>{}c}{Other Ground} & \multicolumn{1}{>{}c}{Structure} & \multicolumn{1}{>{}c}{Vehicle} & \multicolumn{1}{>{}c}{Nature} & \multicolumn{1}{>{}c}{Living being} &  \multicolumn{1}{>{}c}{Object}  \\\midrule\hline
       \specialcell{\textbf{SemanticKITTI} \\ (19 classes)} & \specialcell{Parking \\ Road} & \specialcell{Sidewalk \\ Other-ground} & \specialcell{Building \\ Fence} & \specialcell{Car \\ Truck \\ Bicycle \\ Motorcycle \\ Other-vehicle} & \specialcell{Vegetation \\ Terrain \\Trunk} & \specialcell{Person \\Bicyclist \\Motorcyclist} & \specialcell{Pole \\ Traffic-sign}\\ \hline
        \specialcell{\textbf{KITTI-360} \\ (18 classes)} & \specialcell{Road} & \specialcell{Sidewalk} & \specialcell{Building \\ Wall \\ Fence} & \specialcell{Car \\ Truck \\ Bus \\ Train \\ Motorcycle \\ Bicycle} & \specialcell{Vegetation \\ Terrain} & \specialcell{Person \\ Rider} & \specialcell{Pole \\ Traffic-light \\ Traffic-sign}\\ \hline
        \specialcell{\textbf{nuScenes} \\(16 classes)} & \specialcell{Driveable-surface} & \specialcell{Sidewalk \\ Other-flat} & \specialcell{Manmade \\ Barrier} & \specialcell{Bicycle \\ Bus \\ Car \\ Trailer \\ Truck \\ Construction-vehicle \\ Motorcycle} & \specialcell{Vegetation \\ Terrain} & \specialcell{Pedestrian} & \specialcell{Traffic-cone}\\ \hline
        \specialcell{\textbf{Waymo} \\(22 classes)} & \specialcell{Road \\ Lane-marker} & \specialcell{Sidewalk \\ Walkable \\ Other-ground \\ Curb} & \specialcell{Building} & \specialcell{Car \\ Truck \\ Bus \\ Other-vehicle \\ Bicycle \\ Motorcycle} & \specialcell{Vegetation \\ Tree-trunk} & \specialcell{Pedestrian \\ Bicyclist \\ Motorcyclist} & \specialcell{Traffic-light \\ Construction-cone \\ Pole \\ Sign}\\ \hline
        \specialcell{\textbf{SemanticPOSS} \\ (13 classes)} & \specialcell{Road} & \specialcell{}  & \specialcell{Building \\ Fence} & \specialcell{Car \\ Bike }& \specialcell{Plants \\ Trunk} & \specialcell{People \\ Rider} & \specialcell{Traffic-sign \\ Pole \\ Cone/stone\\ Trash-can}\\ \hline
        \specialcell{\textbf{PandaSet} \\ (36 classes)} & \specialcell{Road \\ Driveway\\ Lane-line-marking \\ Stop-line-marking \\ Other-road-marking} & \specialcell{Ground \\ Sidewalk}& \specialcell{Road-barriers \\ Construction-barrier \\ Building \\ Pylons \\ Other-static}  & \specialcell{Car \\ Bus \\ Pickup-truck \\ Medium-sized-truck \\ Semi-Truck \\ Towed-object \\ Motorcycle \\ Construction-vehicle \\ Uncommon \\ Pedicab \\ Emergency-vehicle \\ Personal-mobility-device \\ Scooter \\ Bicycle \\ Train \\ Trolley \\ Tram} & \specialcell{Vegetation} & \specialcell{Pedestrian \\ Animals} & \specialcell{Signs \\ Cones \\ Construction-signs \\ Rolling-containers}\\ \hline
        \specialcell{\textbf{ParisLuco3D} \\ (45 classes)} & \specialcell{Road \\ Zebra-crosswalk\\ Road-marking\\Parking \\ Bus-lane} & \specialcell{Sidewalk \\Roundabout \\ Bike-lane \\ Central-median} & \specialcell{Building \\ Fence \\ Restaurant-terrace \\ Temporary-barrier \\ Bus-stop \\ Metro-entrance \\ Parking-entrance} & \specialcell{Car \\ Bus \\ Truck \\ Bicycle \\ Motorcycle \\ Construction-vehicle \\ Scooter \\ Trailer } & \specialcell{Vegetation \\ Trunk \\ Terrain \\Vegetation-fence } & \specialcell{Person \\ Bicyclist \\ Motorcyclist} & \specialcell{Pole \\ Ad-spot\\ Traffic-light \\ Garbage-container \\ Garbage-can \\ Traffic-sign \\ Other-object \\Pedestrian-post \\ Light-pole \\ Road-post \\ Bike-rack \\ Bench \\ Bike-post \\ Traffic-cone}\\ \hline\midrule
    \end{tabular}
    
    \caption{Detail of the mapping between the considered datasets and the coarse labels.} 
    \label{tab:coarselabelsex}
\end{table*}
The various labels have self-explanatory names except for Structure and Object. The difference between both stems from nuScenes' object detection dataset, which distinguishes them. Objects correspond to countable and separable objects, such as poles and signs, whereas structures are continuous background elements, such as buildings and barriers.

\begin{figure}[h]
    \centering
    \includegraphics[width=\linewidth]{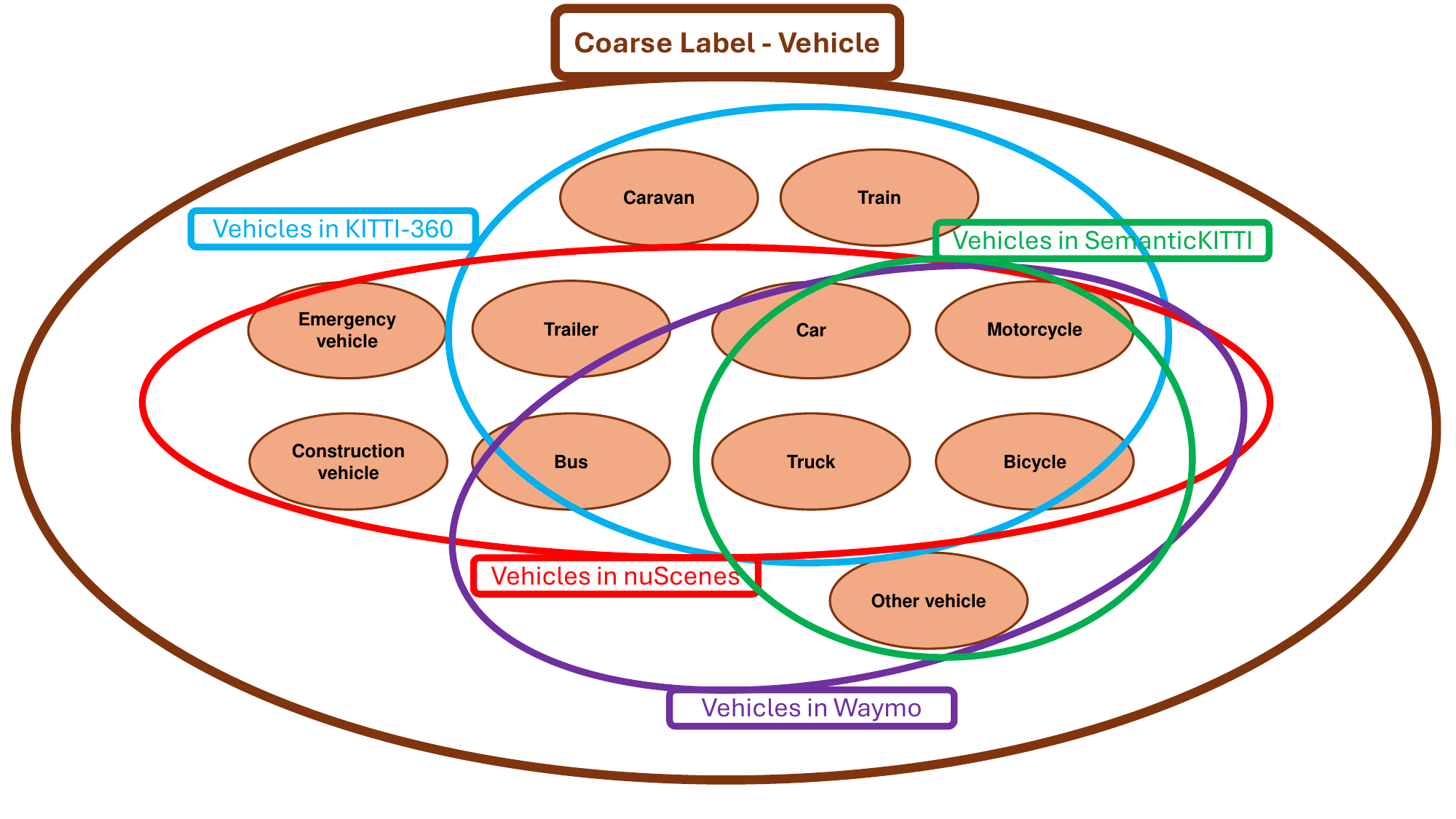}
    \caption{{ Coarse label vehicle and the training labels that constitute it.}}
    \label{fig:Coarse_Vehicle}
\end{figure}

We illustrate the coarse label Vehicle and which fine labels it corresponds to in the training sets in \autoref{fig:Coarse_Vehicle}. This label set is slightly different from the one used in~\cite{cola}, as the difference between dynamic and static objects was deemed ambiguous.

This new label set can be applied to every existing autonomous driving dataset, allowing us to use all the datasets at once to perform multi-source training. Furthermore, as this label set can also be applied to evaluation sets, domain generalization performances can be done on this label set.

\subsection{Our proposed strategies to leverage coarse labels}
As mentioned in the introduction, we propose three new strategies to leverage the coarse labels and the resulting multi-source dataset: COLA-DG, COLA-S2S, and COLA-PT.

COLA-DG investigates leveraging all the datasets at once to enhance generalization performance compared to single-dataset strategies. This strategy is the first tentative attempt at performing multi-source domain generalization in 3D. Alongside the method, an in-depth study is conducted to understand the effect of domain diversity on domain generalization performances.

COLA-DG's limitation is that it can be applied only to coarse labels. To perform fine-level semantic segmentation, which is usually the target task, we propose two strategies: COLA-PT and COLA-S2S.

COLA-PT looks into leveraging the model resulting from COLA-DG, which is robust to domain variation, as a pre-trained model. This constitutes one of the first attempts at supervised pre-training. 

Finally, COLA-S2S combines fine-level and coarse-level annotations to perform semantic segmentation. As highlighted by~\cite{li2022coarse3d}, it is very expensive to perform fine-level annotations. On the contrary, coarse-level annotation is much easier to perform. 
COLA-S2S is a two-step training strategy. First, it performs a pre-training similar to COLA-PT, but it also uses the target set at a pre-training time, assuming the availability of coarse-level annotations. Then, it is finetuned with the fine-level annotations available, which are usually, but not systematically, fewer. 
COLA-S2S is one of the first attempts at multi-source supervised domain adaptation for 3D.

\section{Multi-source domain generalization}
\subsection{Introduction}

In the introduction, we defined multi-source domain generalization as a strategy that uses several different datasets at training time to increase robustness. Furthermore, in Section II, we provided several examples of existing strategies to perform it in 2D and the lack of 3D methods. Finally, in Section III, we described a means of leveraging autonomous driving datasets simultaneously. 

\begin{figure}[h]
    \centering
    \includegraphics[width=0.9\linewidth]{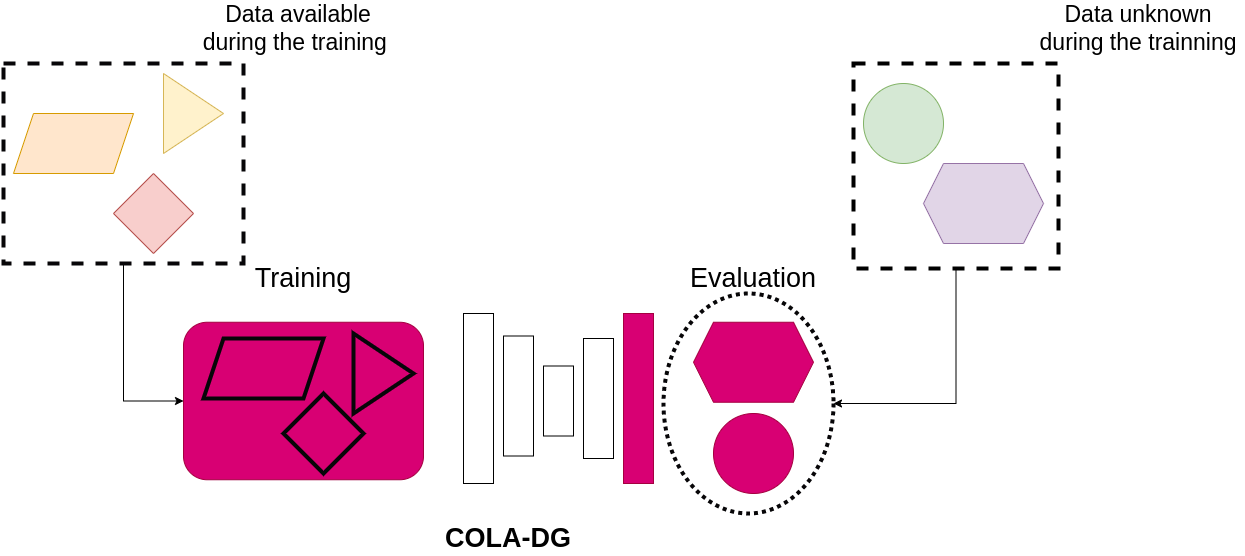}
    \caption{Illustration of COLA-DG. It takes several datasets extracted from various domains as input and remaps them to a single label set through COLA. This new dataset is used to learn a domain-robust semantic segmentation model.}
    \label{fig:COLA_DG}
\end{figure}

In this section, we apply COLA to propose a multi-source domain generalization strategy called COLA-DG for 3D semantic segmentation in autonomous driving. We expect COLA-DG to improve performances compared to single-source learning, as the COLA set we propose displays varied datasets in scenes and acquisition sensors. Furthermore, this dataset is much bigger than any existing single-source dataset. As a result, COLA-DG, unlike single-source domain generalization, can be divided into two components: an increase in the size of the training set and an increase in the diversity of the training set. An illustration of this method can be found in \autoref{fig:COLA_DG}

We are interested in understanding the effect of both of these components separately. Furthermore, we want to understand how diversity improves generalization performance. More specifically, how the various domain shifts are introduced inside the training set is helpful. For this, this section will be divided into three parts: a preliminary study proving the usefulness of multi-source training, a study of COLA-DG for several state-of-the-art models, and an ablation study to understand how multi-source training works.

\subsection{Preliminary study}

In the preliminary study, we are interested in the impact of introducing different sensor resolutions inside the training set. Similar to~\cite{domgen}, Sanchez~\textit{et al.}~\cite{3dlabelprop}, we use new data by lowering the resolution of SemanticKITTI; specifically, we use SemanticKITTI-32 (SK32) and SemanticKITTI-16 (SK16), which emulate 32—and 16-beam LiDAR. These datasets allow us to evaluate the effect of introducing sensor shifts in the training set without incorporating scene and appearance shifts, as we use the exact same sequences.

Following related works~\cite{domgen,segcontrast,3dlabelprop}, we use SRU-Net as our baseline model. We investigate its generalization performance depending on the training set, i.e., trained with only SemanticKITTI and trained with SemanticKITTI and its subsampled versions.

The domain generalization results from this preliminary experiment can be found in \autoref{tab:SK16}. 
\begin{table*}[h]
    \centering
    \begin{tabular}{l||ccc|cccc}
        \multicolumn{1}{c}{Training sets} & SemanticKITTI& SemanticKITTI-16& \multicolumn{1}{c}{SemanticKITTI-32} & SemanticPOSS & ParisLuco3D & Panda64 & PandaFF \\\midrule\hline
        SK & 81.9 & 66.2& 77.7& 35.3& 44.3& 42.8& 42.8\\ \hline
        SK+SK16 & 79.5&73.9 &77.8 &33.5 &52.3 &46.6 & 41.7\\ \hline
        \leftcell{SK+SK16+SK32}  &\textbf{82.5} & \textbf{77.5}& \textbf{80.8}&\textbf{36.9} &\textbf{55.2} &\textbf{50.7 }& \textbf{50.7} \\ \hline \midrule
    \end{tabular}
    \caption{Impact of the introduction of resolution variety on domain generalization performances, results computed with SRU-Net}
    \label{tab:SK16}
\end{table*} 

\begin{table*}[h]
    \centering
    \begin{tabular}{ll||c|cccc}
        \multicolumn{1}{c}{Method}& \multicolumn{1}{c}{Training sets} & \multicolumn{1}{c}{SemanticKITTI} & SemanticPOSS & ParisLuco3D & Panda64 & PandaFF\\\midrule\hline
        KPConv \cite{thomas2019KPConv}& SK&\textbf{78.6} &27.3& 39.7& 25.4& 16.5\\ \hline
        KPConv \cite{thomas2019KPConv}& COLA-DG&73.8 &\textbf{52.3} & \textbf{50.0} & \textbf{56.0} &\textbf{53.4 }\\ \hline\hline
        Cylinder3D \cite{Zhou2020Cylinder3DAE}& SK&81.8 &32.0 &41.0 &29.1 & 7.6\\ \hline
        Cylinder3D \cite{Zhou2020Cylinder3DAE}& COLA-DG&\textbf{82.5} & \textbf{56.6 }& \textbf{67.0} & \textbf{59.8} & \textbf{28.3} \\ \hline\hline
        SRU-Net \cite{mink}& SK&\textbf{81.9} & 35.3& 44.3& 42.8& 42.8\\ \hline
        SRU-Net \cite{mink}&COLA-DG&75.1& \textbf{57.7}& \textbf{59.6}& \textbf{62.5}& \textbf{57.8}\\ \hline\midrule
    \end{tabular}
    \caption{COLA-DG performances compared to single-source training.}
    \label{tab:COLA-DG}
\end{table*}
The first observation is that introducing sensor shift in the training set (SK+SK16) improves domain generalization results, specifically toward lower resolution datasets (SK+SK16$\rightarrow$PL3D) even if the resolution did not appear in the training sensors. Nonetheless, a slight decrease in the high-resolution sensor is observed (SK+SK16$\rightarrow$SK and SK+SK16$\rightarrow$PFF). 

When the number of sensor shifts in the training set (SK+SK16+SK32) is increased, every case improves. The improvement is especially significant for datasets (PL3D) sharing sensor topology with the newly introduced training sets, but it is also seen in unrelated evaluation sets (SK, SP).

Overall, introducing sensor diversity improves domain generalization performance, even for unseen sensors. This preliminary experiment predicts good generalization capabilities for multi-source approaches.

\subsection{COLA for COLA-DG}
We used three semantic segmentation models for the following experiments: KPConv~\cite{thomas2019KPConv}, Cylinder3D~\cite{Zhou2020Cylinder3DAE},  and SRU-Net~\cite{mink}. These are the models currently used by current domain generalization approaches~\cite{domgen,lidog,3dlabelprop,lehner20233d}. In~\cite{3dlabelprop},~\cite{lehner20233d}, it is shown that the naive utilization of the reflectivity channel is detrimental to domain generalization. Based on these observations, we decided to discard the reflectivity channel for all domain generalization experiments.

To assess the usefulness of COLA-DG for domain generalization, we performed two experiments for each model. We trained each model with only SemanticKITTI (SK) and then with the COLA dataset, which is the concatenation of Waymo, SemanticKITTI, KITTI-360, and nuScenes. In both cases, we used the coarse labels for training and evaluation.

Each experiment was done with the same number of epochs (5), as every model was observed to have reached its peak validation performance on the training data by this point. The models were trained by following the parameters found in their respective official repositories: Cylinder3D\footnote{\url{https://github.com/xinge008/Cylinder3D}}, {KPConv\footnote{\url{https://github.com/HuguesTHOMAS/KPConv-PyTorch}}}, {SRU-Net\footnote{\url{https://github.com/mit-han-lab/spvnas}}}.

The quantitative results can be found in \autoref{tab:COLA-DG}.

The first global observation is the systematic improvement in domain generalization when models were trained with COLA, regardless of the model. As mentioned before, this improvement stems from two effects: the size and the domain variety inside the COLA training set.

We want to demonstrate that domain variety is indeed the cause of domain generalization improvement. As dataset size cannot be left out of our analysis, we will dissect both of these effects in Section IV.D. For the remainder of this discussion, we will assume that, indeed, domain variety improves domain generalization performance.

Performance improvements are significant for each domain shift: the significant improvement when evaluating ParisLuco3D and PandaFF proves resilience to sensor shift, and the significant improvement when evaluating SemanticPOSS and Panda64 proves resilience to scene and appearance shifts.

KPConv uses the z-coordinate as an input vector for semantic segmentation. As such, when trained on only one dataset, it is very sensitive to sensor location and displays mediocre domain generalization results. When it is trained on a variety of data, this overfitting pattern is partially erased, which results in drastic improvement (from +11.3\% to +46.9\%). Nonetheless, source-to-source results are slightly reduced.

Cylinder3D results are particularly remarkable. In Sanchez~\textit{et al.}~\cite{3dlabelprop}, it is shown that Cylinder3D is prone to overfitting and is not a very good generalization model. Similar results were obtained in our single-source experiment. However, when Cylinder3D is trained with COLA, it becomes a very strong generalization model, displaying the best source-to-source results alongside the best results on ParisLuco3D. 

These newfound performances can be explained the same way we can explain the overfitting in the single-source case: the pre-processing PointNet. For Cylinder3D, voxelwise features are extracted with a PointNet before the segmentation architecture. While this PointNet overfits when fed only one dataset, it generalizes very well when fed a variety of domains. It is a known result in the registration field~\cite{horache2021mssvconv,poiesi2022learning}.

The very low performance on PandaFF stems from the cylindrical voxelization, which is not parameterized for long-range sensors.

SRU-Net, as already highlighted in~\cite{3dlabelprop}, is the best single-source generalization model. Despite the initially decent performances, multi-source training improves them significantly. Similarly to KPConv, a decrease in source-to-source performance is observed.

Overall, the considered label set is very simple, but we can already observe patterns in the capacity of COLA-DG as a way to tackle domain generalization.

\subsection{Impact of the data variety for COLA-DG}
As mentioned, we want to understand the effect of introducing new datasets on domain generalization performance. In \autoref{tab:ablation_study_COLA-DG}, we incrementally add datasets to understand how variety and size affect generalization.

Here, KITTI-360 plays the role of an expanded SemanticKITTI. Both datasets are acquired in the same location with a similar acquisition setup, but KITTI-360 is much larger. This way, we can assume that there is almost no domain shift between them, and KITTI-360 verifies the effect of dataset size compared to SemanticKITTI.

Using KITTI-360 improves overall generalization results, showing that dataset size impacts performance. However, improvement is nonetheless smaller than when using more diverse training sets.

The results are expected and confirm implicit hypotheses. Adding nuScenes to the training set improves results on ParisLuco3D, as they share the same sensor. Similarly, introducing Waymo in the training set improves results over Panda64. 
 
Using all the datasets together results in a larger improvement than the best improvement when using datasets one at a time. This shows that variety and size interact and that COLA is better than the sum of its parts.

As COLA-based training could be considered overly simple, we compare it with a method referenced in a related work, the multi-head (MH) strategy. For domain generalization, each head is trained with the fine label set of the associated training set. In our case, there are four heads. At inference time, the scores of the heads are remapped to the coarse labels. Then, the scores for each head are added together, and a coarse label can be predicted. 

This inference strategy has two effects: when an inference set is similar to a training set, its associated head will be very confident and provide the final classification. Otherwise, it will be an average of every head. This leads to good performance for sensors that are very close to only one of the training sets, in the case of ParisLuco3D. In the other cases, it performs much worse than COLA.

The generalization performances of COLA-DG compared to MH demonstrate why this method is interesting and efficient.
\begin{table}[t]
    \centering
    \begin{tabular}{l||c|cccc}
        \multicolumn{1}{c}{Training sets} & \multicolumn{1}{c}{SK} & SP & PL3D & P64 & PFF \\\midrule\hline
        SK & \textbf{81.9}&44.3& 35.3& 42.8& 42.8 \\ \hline
        K360 &61.8 &42.5 & 50.1& 44.2& 42.7 \\ \hline
        K360 + SK  & 70.7&43.2 &48.5 &49.9 &47.4 \\ \hline
        K360 + NS  & 61.0&46.3 &53.5 &48.7 &39.4 \\ \hline
        K360 + W  & 63.1&54.3 &52.4 &59.5 &52.2\\ \hline
        K360 + SK + W + NS & 75.1&\textbf{57.7}& 59.6& \textbf{62.5}& \textbf{57.8}\\ \hline
        K360 + SK + W + NS (MH)& 59.9&54.2 &\textbf{62.6} &49.4 &30.3\\ \hline \midrule
    \end{tabular}
    \caption{Impact of data variety on domain generalization performance; results computed with SRU-Net.}
    \label{tab:ablation_study_COLA-DG}
\end{table} 

\subsection{Benchmark multi-source domain generalization}

In addition to the analysis presented here, we expand the ParisLuco3D~\cite{sanchez2023parisluco3d} online benchmark to include a track with the coarse labels. This new track will allow people to compare multi-source domain generalization methods on a unified label set and is available following the link: \url{https://npm3d.fr/parisluco3d}.

\section{Multi-source LiDAR semantic segmentation}
\subsection{Intuition}
In the previous section, we restricted the discussion of one result in \autoref{tab:COLA-DG}. When trained with COLA, Cylinder3D improves its source-to-source segmentation performance. This result suggests that multi-source training could be used to perform LSS. We call this task multi-source LiDAR semantic segmentation.  It is a supervised domain adaptation strategy.

\begin{figure}[t]
    \centering
    \includegraphics[width=0.9\linewidth]{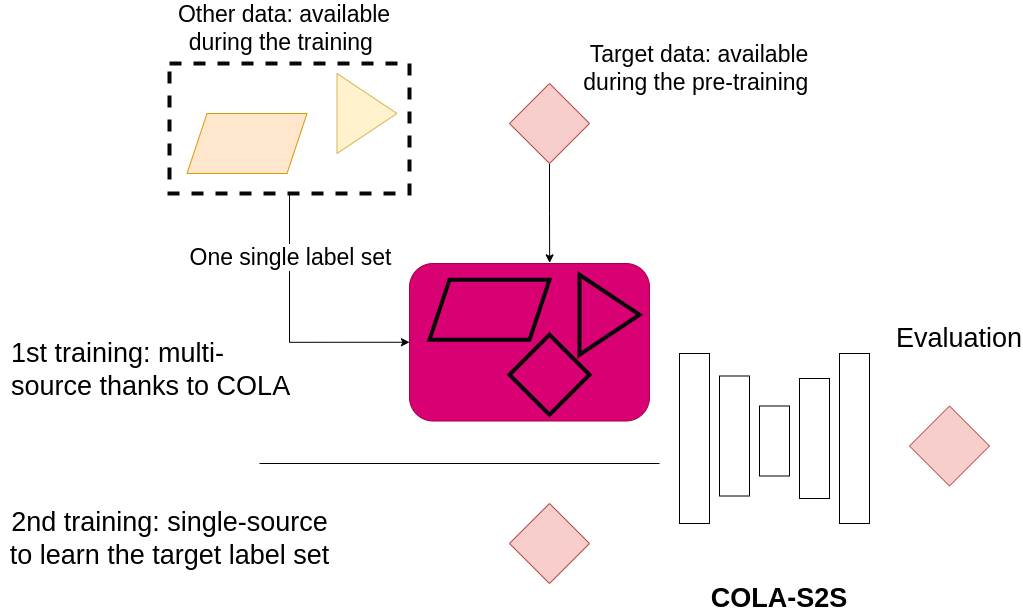}
    \caption{Illustration of COLA-S2S. The first step takes several datasets extracted from various domains as input and remaps them to a single label set through COLA. This new dataset, which includes the target set, is used to learn a domain-robust semantic segmentation model. This model is then refined with the target dataset to infer the right label set.}
    \label{fig:cola_s2s}
\end{figure}

In this task, we expect examples of the evaluation set to be available, at least partially, during training, as they would be for typical LSS. Based on this hypothesis, we design a pipeline called COLA-S2S that can allow us to leverage multi-source information for source-to-source segmentation. COLA-S2S is illustrated in~\autoref{fig:cola_s2s}.

The difference between COLA-DG and COLA-S2S is that COLA-S2S requires the inference labels to be the same as the fine label set of the target, which means COLA cannot be applied throughout the pipeline. We propose a two-step training process. First, following the COLA-DG training process, the COLA-S2S model is trained with a concatenation of datasets, including the target set, supervised by the coarse labels. This step allows the model to learn useful representation of the scene and disambiguates the main road components. Then, the COLA-S2S model is \textit{finetuned} with the target set only, supervised by its fine labels.

The remainder of this section describes the experimental protocols more precisely and presents the results.

\subsection{Experiments}

We set up two experiments. As COLA-S2S is expected to leverage large amounts of data, we use nuScenes (NS) and SemanticKITTI (SK) as target sets to ensure that standard LSS methods can still learn and that we can evaluate the usefulness of COLA-S2S. All experiments were performed with SRU-Net and Cylinder3D. SRU-Net is the baseline for LSS, and Cylinder3D is reaching state-of-the-art performances.

Inspired by related works~\cite{segcontrast,nunes2023temporal}, we emulate two cases: one where the target set is fully finely annotated and one where the target set is partially finely annotated and fully coarsely annotated. For the second case, the coarse labels are known for all the data, but only a small part of the data has the target label annotations.

Following the TARL~\cite{nunes2023temporal} protocol\footnote{\url{https://github.com/PRBonn/TARL}}, we trained our SRU-Net models with the same number of epochs and the same learning rate as the TARL authors propose for finetuning. For the first training step, we re-employ the model trained for COLA-DG, which does not use reflectivity. We add reflectivity to our data for the second step to ensure the best performances. 

In the case of low fine annotation availability, we propose to add a baseline that uses only coarsely annotated SemanticKITTI as the pre-training set. This way, we can validate the usefulness of coarse-level annotation for single-source training and assess the improvement made by using COLA-S2S.

{ We compare our approach with SegContrast~\cite{segcontrast} and TARL~\cite{nunes2023temporal}, knowing that both methods are unsupervised strategies. Multisource pre-training and, more broadly, supervised pre-training are nonexistent in 3D, so existing pre-training methods are all unsupervised. The objective of COLA-S2S is to highlight a novel credible alternative to unsupervised pre-training. Due to availability, the comparison with SegContrast~\cite{segcontrast} and TARL~\cite{nunes2023temporal} was conducted exclusively using the SRU-Net model.}

\subsection{Results}

Results for SRU-Net can be found in \autoref{tab:COLA_SKNS} and \autoref{tab:COLA_SK_percent}, and results for Cylinder3D in \autoref{tab:COLA_SKNS_C3D} and \autoref{tab:COLA_SK_percent_C3D}.

The COLA-S2S results for SRU-Net on full data availability are particularly good, showing a significant improvement for both SemanticKITTI (+5.3\%) and nuScenes (+4.0\%) over the standard LSS method. Furthermore, COLA-S2S is a more meaningful strategy than geometry-based data exploitation as it outperforms SegContrast and TARL. This result highlights the importance of data diversity for LSS.

Regarding the low availability experiment, COLA-S2S proves useful, being on par with the state-of-the-art method, TARL. Nonetheless, we again demonstrate the benefit of using a multi-source rather than a single-source dataset, as COLA-S2S systematically outperforms using only coarsely annotated SemanticKITTI for training.
\begin{table}[t]
    \centering
    \begin{tabular}{l||cc}
        \multicolumn{1}{c}{Method} & SK & NS  \\ \midrule \hline
        No pre-training  & 59.6 & 66.0\\ \hline
        SegContrast \cite{segcontrast}& 60.5 & 67.7\\ \hline
        TARL \cite{nunes2023temporal} & 61.5 & 68.3\\ \hline
        COLA-S2S& \textbf{64.9}  & \textbf{70.0}\\ \hline \midrule
    \end{tabular}
    \caption{COLA-S2S results on nuScenes and SemanticKITTI; computed with SRU-Net.}
    \label{tab:COLA_SKNS}
\end{table} 

\begin{table}[t]
    \centering
    \begin{tabular}{l||ccccc}
        \multicolumn{1}{c}{Method} & 0.1\% & 1 \% & 10\% & 50\% & 100\%  \\ \midrule \hline
        No pre-training & 25.6 &41.7 &53.9 &58.3 &59.6 \\ \hline
        SK & 36.3 & 50.2& 59.5 &\textbf{62.8} & 62.5 \\ \hline
        SegContrast \cite{segcontrast}& 34.8 & 47.4& 55.2& 58.3&60.5\\ \hline
        TARL \cite{nunes2023temporal}& \textbf{38.6} & 51.4 & \textbf{60.3} & 61.4 & 61.5 \\ \hline
        COLA-S2S& 38.3 & \textbf{51.8}& 58.0 & 60.8 &  \textbf{64.9} \\ \hline \midrule
    \end{tabular}
    \caption{COLA-S2S results on SemanticKITTI, depending on the number of fine labels available, computed with SRU-Net.}
    \label{tab:COLA_SK_percent}
\end{table} 
COLA-S2S results for Cylinder3D are mixed. On full data availability, COLA-S2S provides minor improvement for nuScenes (+0.4\%) and none for SemanticKITTI. In the case of low availability, COLA-S2S improves results over using only SemanticKITTI for training, similar to the SRU-Net case.

Overall, COLA-S2S is an interesting strategy that can improve results for a very small cost, as it leverages coarse annotations. It is specifically useful for SRU-Net, allowing it to perform better than Cylinder3D thanks to COLA-S2S despite having 4\% less mIoU (mean Intersection-over-Union) with a standard approach.

\begin{table}[t]
    \centering
    \begin{tabular}{l||cc}
        \multicolumn{1}{c}{Method} & SK & NS  \\ \midrule \hline
        No pre-training  & \textbf{63.7} & 74.8\\ \hline
        COLA-S2S& 63.0 & \textbf{75.2} \\ \hline \midrule
    \end{tabular}
    \caption{COLA-S2S results on nuScenes and SemanticKITTI; computed with Cylinder3D.}
    \label{tab:COLA_SKNS_C3D}
\end{table}

\begin{table}[t]
    \centering
    \begin{tabular}{l||ccccc}
        \multicolumn{1}{c}{Method} & 0.1\% & 1 \% & 10\% & 50\% & 100\%  \\ \midrule \hline
        No pre-training  & 34.4 & 48.5 & 58.2&61.1 &\textbf{63.7}  \\ \hline
        SK & 43.8 & 51.3 & 58.0 & 61.4 & 61.3\\ \hline
        COLA-S2S& \textbf{46.5} & \textbf{53.7}& \textbf{60.5}& \textbf{62.2} & 63.0\\ \hline \midrule
    \end{tabular}
    \caption{COLA-S2S results on SemanticKITTI, depending on the amount of fine labels available; computed with Cylinder3D.}
    \label{tab:COLA_SK_percent_C3D}
\end{table}

\section{Pre-training for LiDAR semantic segmentation}
\subsection{Intuition}

\begin{figure}[h]
    \centering
    \includegraphics[width=0.9\linewidth]{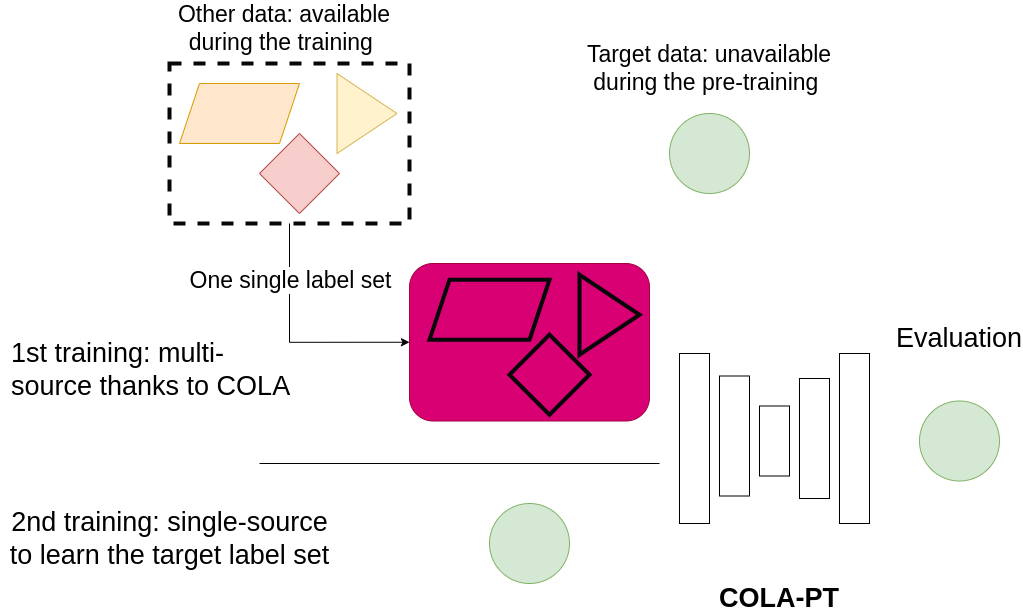}
    \caption{Illustration of COLA-PT. The first step takes several datasets extracted from various domains as input and remaps them to a single label set through COLA. This new dataset is used to learn a domain-robust semantic segmentation model. This model is then refined with the target dataset to infer the right label set.}
    \label{fig:cola_pt}
\end{figure}

Based on great COLA-DG results and encouraging COLA-S2S results, the next step would be to consider multi-source pre-training. As a reminder, the objective of pre-training neural networks is to learn reusable geometric representations that can be easily leveraged during finetuning to enhance final performance or to require less finetuning data. Here, the pre-training and finetuning tasks are the same, namely semantic segmentation. Based on COLA-DG results, we have a guarantee that learned representations are useful in new unseen data. Our proposed multi-source approach to pre-training is called COLA-PT and is illustrated in~\autoref{fig:cola_pt}.

Contrary to COLA-S2S, COLA-PT expects fine-tuning data to be unavailable during pre-training. The training pipeline is as follows: the COLA-PT model is pre-trained with COLA and then fine-tuned with the target data supervised by its fine label set.

In this section, we introduce our experiments, propose a qualitative analysis of why COLA can be a useful pre-training method, and present the quantitative results.

\subsection{Experiments}

We follow a very similar protocol to the one for COLA-S2S. The main difference is the target set. Here it is SemanticPOSS. Despite using SemanticKITTI as the target in the literature, we do not use it here, as the resulting pre-training set was deemed too small. Indeed, for fairness of evaluation, when SemanticKITTI is the target set, KITTI-360 would need to be removed from the pre-training set, resulting in a much smaller dataset. We are not subject to this issue when using SemanticPOSS as the target set.

We study the finetuning results on the full SemanticPOSS dataset and various subsets of annotated frames. We introduce a baseline supervised pre-training method that uses only SemanticKITTI for pre-training. This way, we can gauge the impact of COLA-PT relative to simple supervised pre-training. We perform these experiments with SRU-Net and Cylinder3D. Furthermore, unlike previous sections, we perform these experiments with and without the reflectivity channel. While it made sense not to use it for generalization and for source-to-source segmentation, here, for pre-training, it was not clear which one was better. Indeed, following COLA-DG, the pre-trained model is trained without reflectivity but, contrary to COLA-S2S, the finetuning dataset is small, and it is unsure whether the model will transition smoothly from data without reflectivity for pre-training and with reflectivity for finetuning.

{ We compare our results with SegContrast~\cite{segcontrast}, the only method that performed this experiment. It is important to remind that SegContrast is an unsupervised single source pre-training method, whereas COLA-S2S employs a supervised multi-source pre-training strategy.} Following the SegContrast protocol *, we trained our SRU-Net models with the same number of epochs and the same learning rate as the one the SegContrast authors propose for fine-tuning.

In their paper, the SegContrast authors concluded that geometric unsupervised pre-training performs better than supervised pre-training—their results back their claims. Our work differs from theirs in that we leverage a larger amount of available annotated data. 

\begin{figure*}[h]
    \centering
    \subfloat[SemanticKITTI]{
        \includegraphics[width=.31\linewidth]{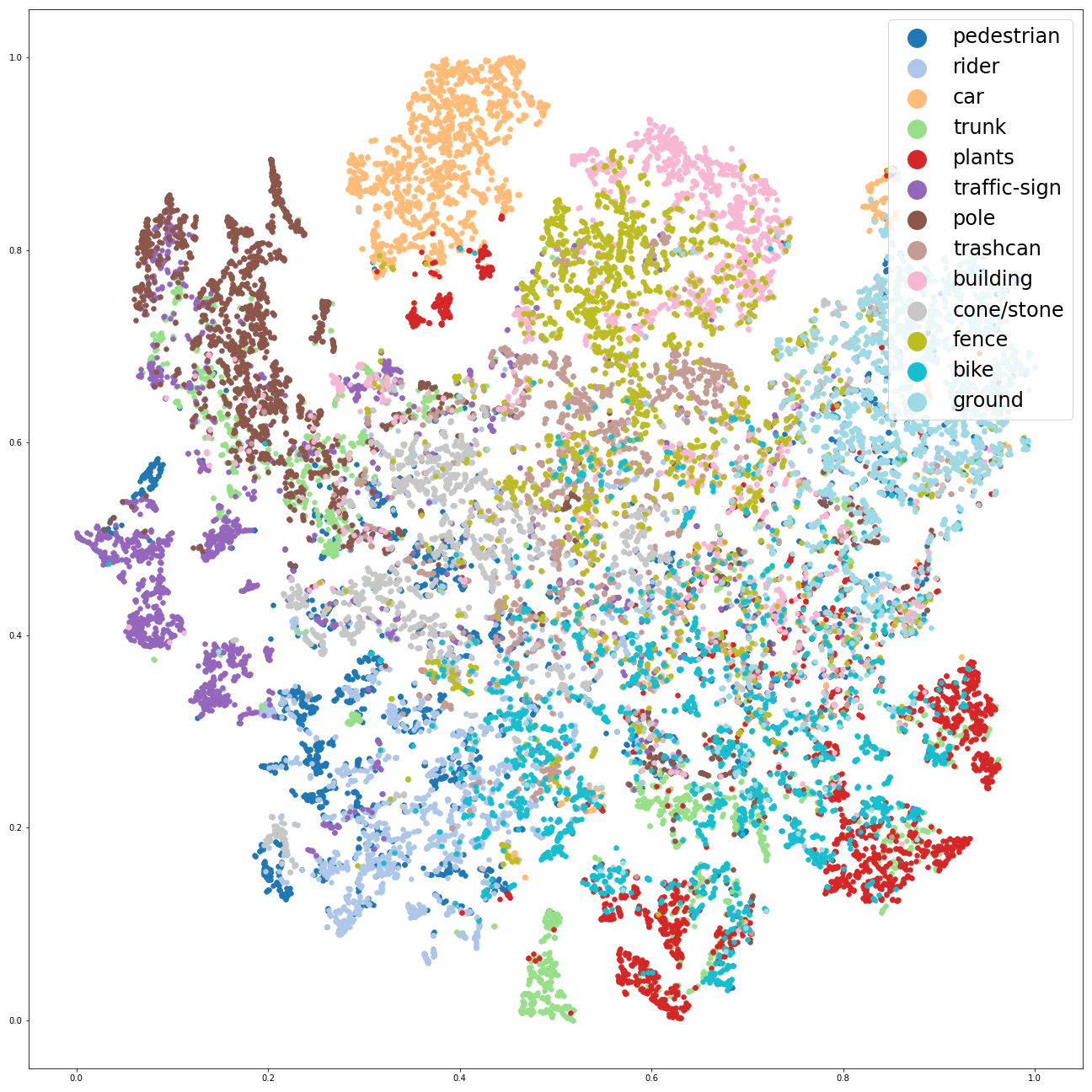}
    }
    \subfloat[COLA-PT]{
        \includegraphics[width=.31\linewidth]{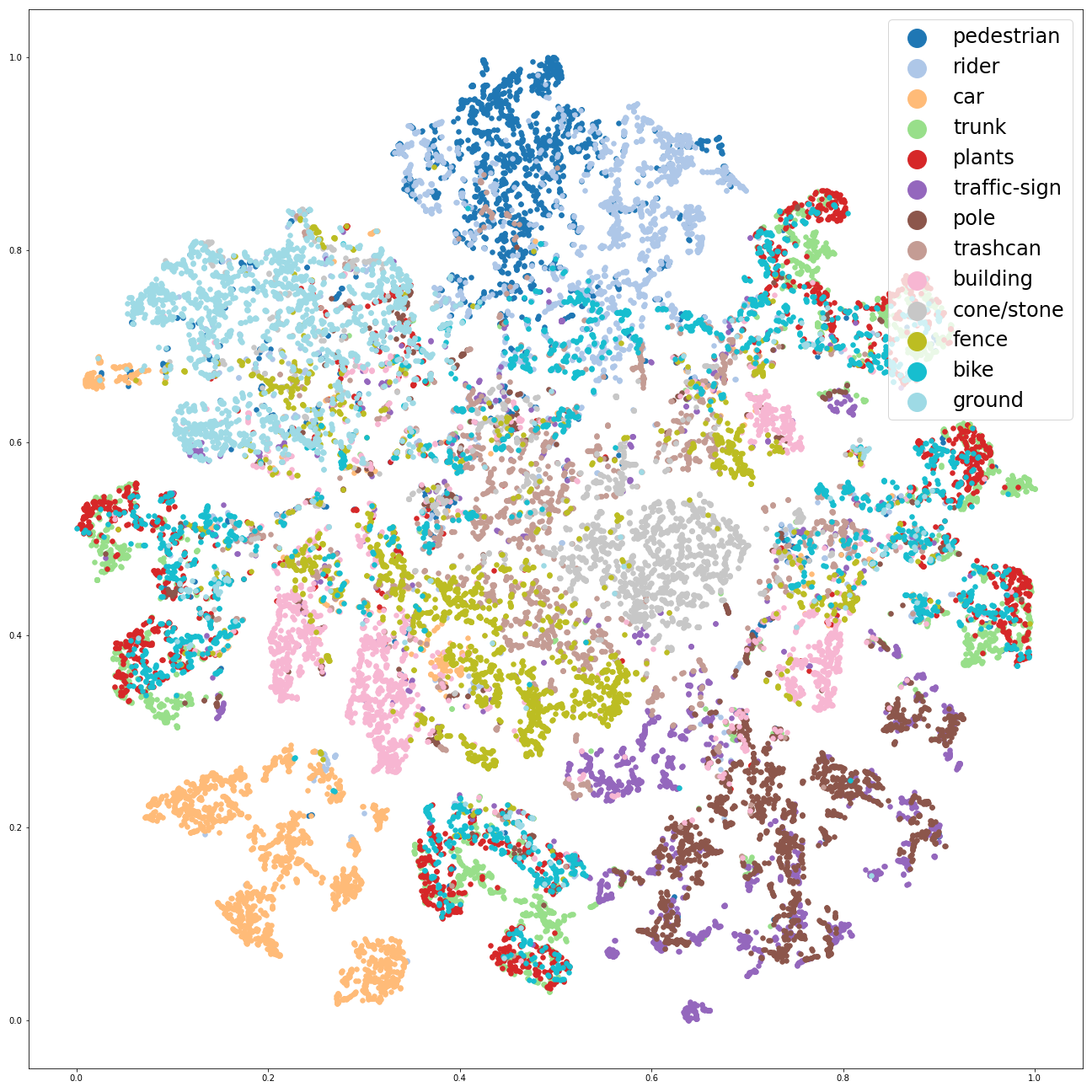}
    }
    \subfloat[SegContrast]{
        \includegraphics[width=.31\linewidth]{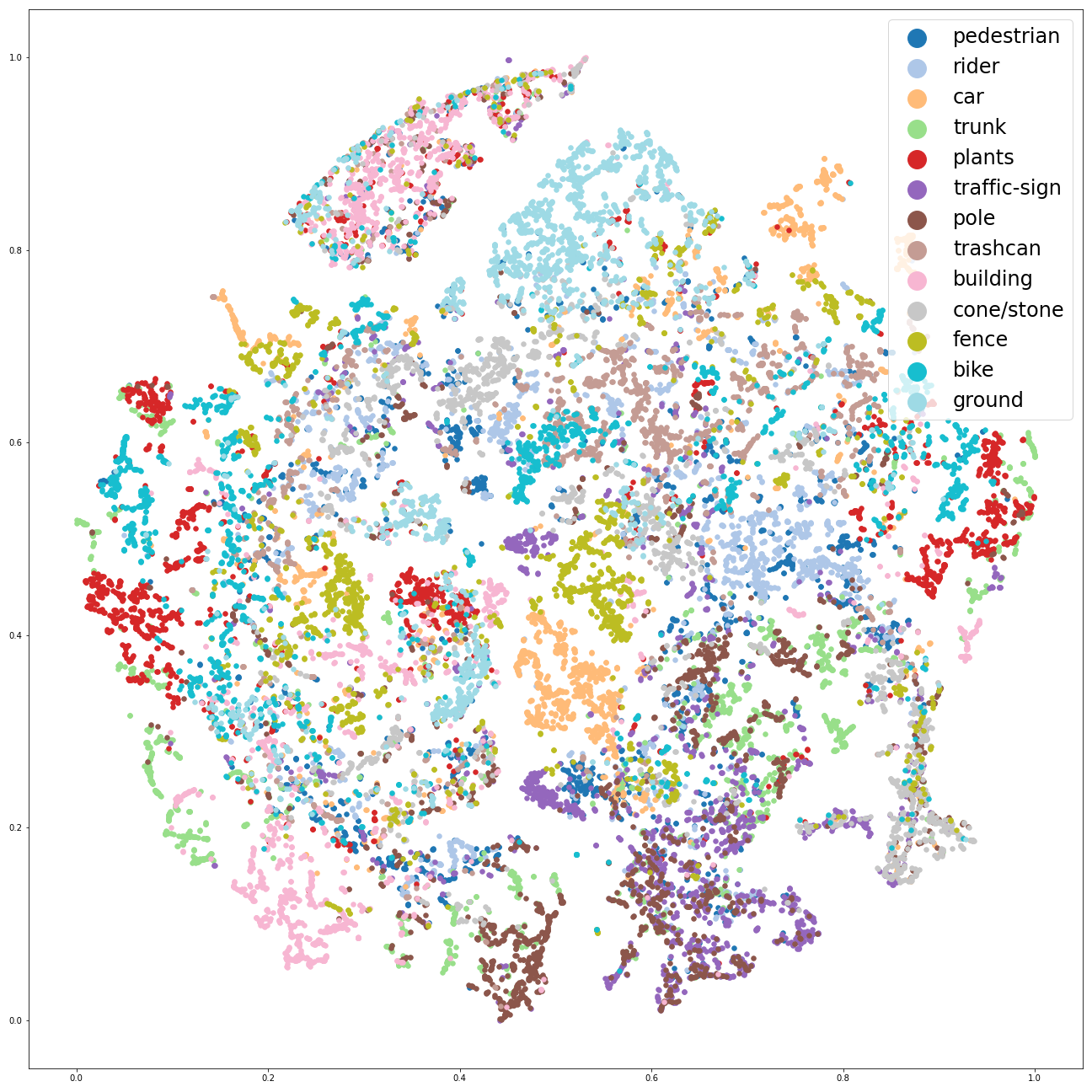}
    } 
    \caption{t-SNE analysis of the feature extracted by an SRU-Net depending on the pre-training strategy. From left to right: SemanticKITTI, COLA-PT, SegContrast.}
    \label{fig:tsne}
\end{figure*}

\subsection{Qualitative results}

We first perform a qualitative analysis to highlight our pre-training's generalization power and show its meaningfulness compared to geometric pre-training. For this, we do a t-SNE analysis of extracted features from the target set before finetuning. This way, we can see if some information is already in the embedded space, even though the target set was never seen by the pre-trained model. We then color each point by its associated label. These graphs can be seen in \autoref{fig:tsne}.

We expect supervised pre-training to semantically pre-cluster points in the t-SNE space, as they have already learned coarse semantic information, whereas SegContrast was only exposed to geometric information.

We observed that SegContrast is confused between classes of similar geometry or with similar neighbors, such as Traffic-sign and Pole. Inversely, supervised pre-training can help disambiguate classes based on their semantics, such as car, building, and fence. Nonetheless, the impact of the supervised pre-training label set can be observed as there is confusion between Rider and Pedestrian, which belong to the same coarse label.

Classes that are very distinctive geometrically and semantically, such as ground, are well recognized by all types of pre-training.

Qualitatively, we see that classes seem more understood by semantic pre-training, proving its usefulness.

\subsection{Results}

Even though the qualitative analysis suggests that understanding SemanticPOSS is much better at the end of the supervised pre-training, the quantitative results are not as clear-cut.

For SRU-Net (\autoref{tab:COLA_SP}), COLA-PT outperforms the single-source supervised pre-training systematically.

Then, compared to SegContrast, COLA-PT is more efficient for very low amounts of available data (0.01\%) and is on par in other cases, which means it achieves state-of-the-art performances. We observed that despite the pre-trained model not using reflectivity, it is more valuable to use it for fine-tuning.

\begin{table}[t]
    \centering
    \begin{tabular}{l||ccccc}
        \multicolumn{1}{c}{Method} & 0.1\% & 1 \% & 10\% & 50\% & 100\%  \\\midrule \hline
        No pre-training & 33.1 &43.1 &57.3 &63.3 &64.2 \\ \hline
        SK & 42.9 &54.4 & 60.2& 64.3& 64.5\\ \hline
        SegContrast \cite{segcontrast}& 43.7 & \textbf{55.2}& 60.3& \textbf{64.6}& \textbf{64.9}\\ \hline
        COLA-PT w/o r& 44.5  & 53.1 & 58.8 & 62.0 & 63.3  \\ \hline
        COLA-PT& \textbf{45.1} & 54.5 & \textbf{60.6} & 63.6 &  \textbf{64.9} \\ \hline \midrule
    \end{tabular}
    \caption{Finetuning results on SemanticPOSS, depending on the number of labels available, computed on SRU-Net.}
    \label{tab:COLA_SP}
\end{table} 

On Cylinder3D (\autoref{tab:COLA__C3D_SP}), results differ slightly from those of SRU-Net. While COLA-PT results in the best outcome in most cases, it is unclear if it is useful to use reflectivity or not as an input channel. For a very small amount of data (0.1\%), COLA-PT does not bring an improvement, which is surprising.

\begin{table}[t]
    \centering
    \begin{tabular}{l||ccccc}
        \multicolumn{1}{c}{Method} & 0.1\% & 1 \% & 10\% & 50\% & 100\%  \\\midrule \hline
        No pre-training & \textbf{39.6} & 52.1 & 59.0& 64.3& 65.0 \\ \hline
        SK &  38.4 & 52.4 & \textbf{60.3}& 64.6& 65.6\\ \hline
        COLA-PT w/o r& 35.7 & \textbf{52.9} & 58.3& 64.6 & \textbf{65.9} \\ \hline
        COLA-PT& 38.7 & 51.1 & 60.1 &  \textbf{65.2} & 65.5\\ \hline \midrule
    \end{tabular}
    \caption{Finetuning results on SemanticPOSS, depending on the number of labels available, computed on Cylinder3D.}
    \label{tab:COLA__C3D_SP}
\end{table} 

Overall, COLA-PT is a strong choice of semantic segmentation pre-training, and it achieves state-of-the-art performances in several cases. 

\subsection{Influence of the COLA-PT label set}

To conclude on COLA-PT, we would like to discuss the effect that the pre-training label set has on fine-tuning performances. Because the pre-trained model is not concerned with evaluation, methods such as intersection—and union-based training can be used. 

As such, we study six different approaches for labeling of the pre-training set in two different cases: 0.1\% available label, which is the hardest learning task, and 100\% available label.

These six different strategies are as follows:
\begin{itemize}
     \item COLA: our coarse labels method (seven labels), 
     \item COLA-5: a reduction to five labels of COLA (\textit{Ground}, \textit{Vehicle}, \textit{Manmade}, \textit{Pedestrian}, \textit{Vegetation}), 
     \item COLA-9: an extension to nine labels of COLA (\textit{Driveable Ground}, \textit{Structure}, \textit{4-Wheeled}, \textit{Nature}, \textit{Pedestrian}, \textit{Object}, \textit{Other Ground}, \textit{Pole\&Sign}, \textit{2-Wheeled}),
     \item Intersection: Keep the nine labels identified as part of the intersection between the label sets,
     \item Union: use the 32 different labels that compose the union of four training label sets,
     \item MH: the multi-head method. 
\end{itemize}

The results are shown in \autoref{tab:COLA_SP_label}. The finer the pre-training label set is, the better it performs for very low amounts of data available. Conversely, the coarser the label, the better it performs at high amounts of available data.

It must be noted that COLA strategies are independent of the original label sets, whereas Union, Intersection, and MH are not.

\begin{table}[t]
    \centering
    \begin{tabular}{lc||cc}
        \multicolumn{1}{c}{Method} & \multicolumn{1}{c}{\# labels} & 0.1\% & 100\%  \\ \midrule \hline
        No pre-training & N/A& 33.1 & 64.2 \\ \hline
        COLA & 7 & 44.5 &  64.9\\ \hline
        COLA-5 & 5&43.7 &  \textbf{65.0}\\ \hline
        COLA-9 & 9&45.2 &  64.8\\ \hline
        Intersection & 9&43.3 & 64.6\\ \hline
        Union & 32&\textbf{46.6} & 63.6  \\ \hline
        MH & 59&45.3 & 64.6 \\ \hline \midrule

    \end{tabular}
    \caption{Label set choice influences SemanticPOSS finetuning results; computed on SRU-Net.}
    \label{tab:COLA_SP_label}
\end{table}

\section{Conclusion and limitations}

We have introduced a novel relabelling strategy called COLA that can be used to perform multi-source training for almost no human cost. This way, to the best of our knowledge, we performed the first multi-source experiments for LiDAR semantic segmentation in autonomous driving.

We explored the usefulness of multi-source training to improve domain generalization, source-to-source segmentation, and pre-training performances. We achieved systematic improvements and even demonstrated that these results were robust relative to the label set chosen.

More precisely, multi-source strategies are more robust to domain shift, as demonstrated in \autoref{tab:COLA-DG}. However, while previous multi-source approaches in 2D computer vision were mainly used to improve robustness, we also demonstrated that multi-source approaches are useful for single-domain evaluation, {as seen in~\autoref{tab:COLA_SKNS_C3D}}. Finally, multi-source training results in a model with strong priors, which can be used for pre-training { as shown by the results in}~\autoref{tab:COLA_SP}.

As a result, we recommend using multi-source systematically as long as training time is not an issue. Nonetheless, { we believe that there} is still much work to be done to explore and fully understand multi-source training.

Nonetheless, the proposed strategies have several limitations. First, they assume the availability of coarse labels at training time. While this annotation type is less costly than typical annotations, they are not systematically available. Second, training time and needed resources are increased, which goes against the flow of current research that looks into label- and resource-efficient methods.

We believe multi-source training is a credible alternative to single-source strategies in the abovementioned fields.

{\small
\bibliographystyle{IEEEtran}
\bibliography{bib}
}
\end{document}